\documentclass[sigconf]{acmart}

\usepackage{balance} 
\usepackage{enumitem}

\renewcommand\footnotetextcopyrightpermission[1]{}

\AtBeginDocument{%
  }

\acmDOI{}
\acmISBN{}
\acmConference[Preprint]{Preprint Version}{December 2025}{Liverpool, UK}

\acmYear{}
\copyrightyear{}

\begin{document}

\title{Rethinking Multi-Agent Intelligence Through the Lens of Small-World Networks}

\author{Boxuan Wang, Zhuoyun Li, Xiaowei Huang, Yi Dong$^{*}$}


\affiliation{%
  \institution{School of Computer Science and Informatics,  University of Liverpool}
  \city{Liverpool}
  \country{United Kingdom}
}

\authornote{Corresponding Author:  Yi.Dong@liverpool.ac.uk \\Preprint, Under Review}









\renewcommand{\shortauthors}{Wang et al.}


\begin{abstract}
Large language models (LLMs) have enabled multi-agent systems (MAS) in which multiple agents argue, critique, and coordinate to solve complex tasks, making \emph{communication topology} a first-class design choice. Yet most existing LLM-based MAS either adopt fully connected graphs, simple sparse rings, or ad-hoc dynamic selection, with little structural guidance. In this work, we revisit classic theory on small-world (SW) networks and ask: \emph{what changes if we treat SW connectivity as a design prior for MAS?} We first bridge insights from neuroscience and complex networks to MAS, highlighting how SW structures balance local clustering and long-range integration. Using multi-agent debate (MAD) as a controlled testbed, experiment results show that SW connectivity yields nearly the same accuracy and token cost, while substantially stabilizing consensus trajectories. Building on this, we introduce an uncertainty-guided rewiring scheme for scaling MAS, where long-range shortcuts are added between epistemically divergent agents using LLM-oriented uncertainty signals (e.g., semantic entropy). This yields \emph{controllable} SW structures that adapt to task difficulty and agent heterogeneity. Finally, we discuss broader implications of SW priors for MAS design, framing them as stabilizers of reasoning, enhancers of robustness, scalable coordinators, and inductive biases for emergent cognitive roles.
\end{abstract}



\keywords{Multi-Agent Systems, Small-World Network, Multi-Agent Debate, Large Language Models}

\maketitle

\section{Introduction}

Large language models (LLMs) have enabled a new generation of multi-agent systems (MAS) in which multiple LLM agents interact, critique, and collaborate to solve complex tasks. Frameworks such as AutoGen~\citep{wu2024autogen} and CAMEL~\citep{li2023camel} demonstrate that organizing agents into conversational roles can enhance reasoning, while benchmarks like AgentBench~\citep{liu2024agentbench} reveal persistent limitations in long-horizon planning, coordination, and self-correction. These observations underscore that the structure of inter-agent communication is becoming a central design axis for multi-agent intelligence.

A growing line of work investigates how communication \emph{topology} affects collective reasoning. In multi-agent debate (MAD), a widely studied testbed for collaborative reasoning, agents iteratively exchange arguments and refine their answers~\citep{du2023improving,liang-etal-2024-encouraging,kaesberg-etal-2025-voting,10.1145/3726302.3730092}. Although early MAD implementations commonly adopted a fully-connected topology, recent work challenges this assumption. Li et al.~\citep{li2024sparse} show that even a simple ring topology, where each agent communicates only with its neighbors, matches or surpasses fully-connected debate while using far fewer tokens, suggesting that unrestricted information sharing may introduce error.

\begin{figure}[t]
    \centering
    \includegraphics[width=0.49\textwidth]{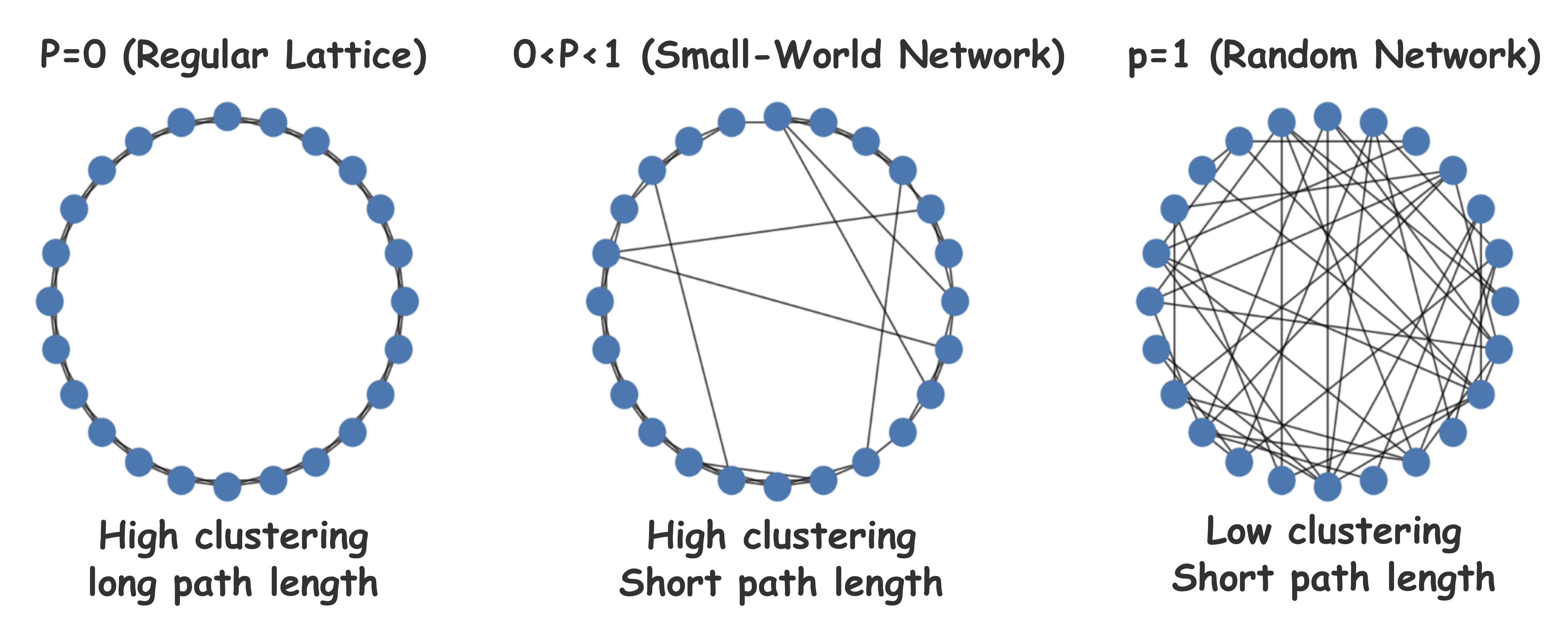}
    \caption{Networks under different rewiring probabilities \(p\), illustrating the transition from regular lattice (\(p=0\)) to small-world (\(0<p<1\)) to random network (\(p=1\)).}
    \label{fig:intro}

\end{figure}

These findings raise a broader question: what kinds of communication topologies fundamentally support efficient multi-agent reasoning? Fully-connected and ring structures are commonly used topologies in MAS, yet natural and social systems often adopt a different organizational principle: small-world (SW) structures, which were originally formalized by \citet{watts1998collective}. The SW network is characterized by high local clustering and short average path lengths between nodes, as illustrated in Figure \ref{fig:intro}. A SW structure combines high local clustering with a short average path length, typically achieved by introducing a small number of long-range ``shortcut'' edges. Such structures are known to facilitate rapid information flow and robust collective behavior by cutting average path length in biological neural systems \cite{bullmore_complex_2009,sporns_small-world_2004,mohammadi2024graph}.

Despite their ubiquity in brain science and other subjects, SW structures have been surprisingly absent from current LLM-based MAS research. Existing work either fixes a dense topology, restricts communication to a ring, namely sparse communication ~\citep{li2024sparse}, or dynamically selects agents without explicit structural priors~\citep{liu2024dylan}. Even recent analyses suggesting that LLM agents may spontaneously form SW-like patterns~\citep{papachristou2024network} have not been translated into concrete MAS design. This gap motivates a systematic investigation: \emph{could deliberately engineered SW structures improve the stability and efficiency of LLM-based MAS reasoning and collaboration?}

In this paper, we take a first step toward addressing this question. We argue that SW communication structures represent a promising yet underexplored design dimension for future LLM-based MAS. We advocate for a structural research agenda centered on the principles of SW, grounded in the following two perspectives:

\begin{itemize}[left=5pt]
\item \textbf{SW as a reasoning stabilizer in MAS:} We introduce SW topologies into MAD and carried out pioneering preliminary experiments. We find that inserting a small number of shortcut edges improves consensus stability, maintains competitive accuracy, and reduces token cost. This suggests SW structures offer an inductive bias for robust collaborative reasoning, which can be a promising research direction.

\item \textbf{SW as a controllable architecture for scalable MAS:} we outline a blueprint for constructing tunable SW communication graphs in general MAS environments. By leveraging methods of uncertainty quantification (e.g., semantic entropy), we can modulate the degree of connectivity to enhance the stability of consensus formation in complex systems.
\end{itemize}

\begin{figure}[t]
    \centering
    \includegraphics[width=0.49\textwidth]{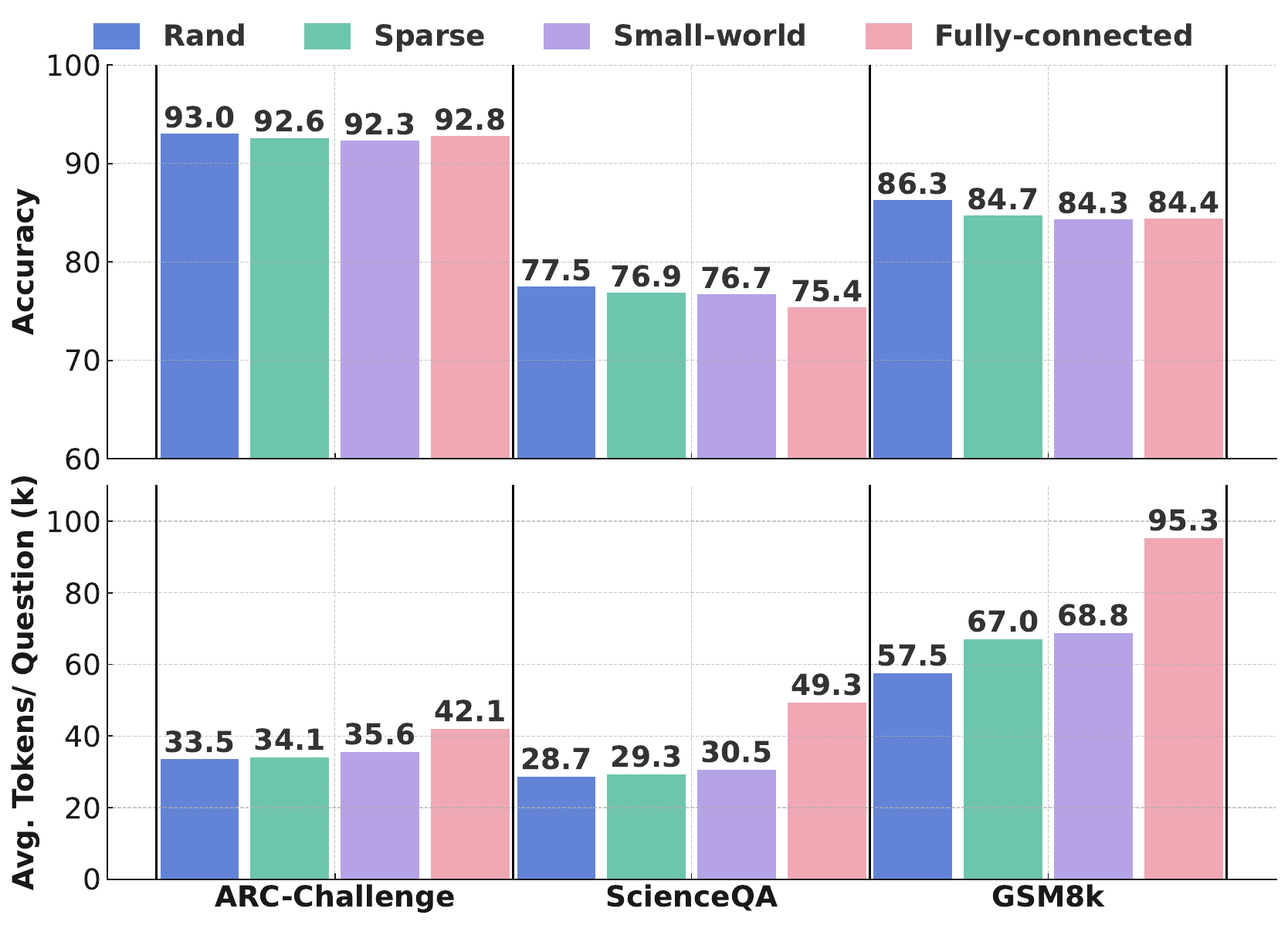}
    \caption{
    Performance of different communication topologies in multi agent debate. The top panel reports accuracy across three datasets.  
    The bottom panel reports the average communication cost measured in tokens per question. The experiment is performed using GPT-4o-mini with 8 agents.}
    \label{fig:mad_topologies}
\end{figure}
\begin{figure}[t]
    \centering
    \includegraphics[width=1\linewidth]{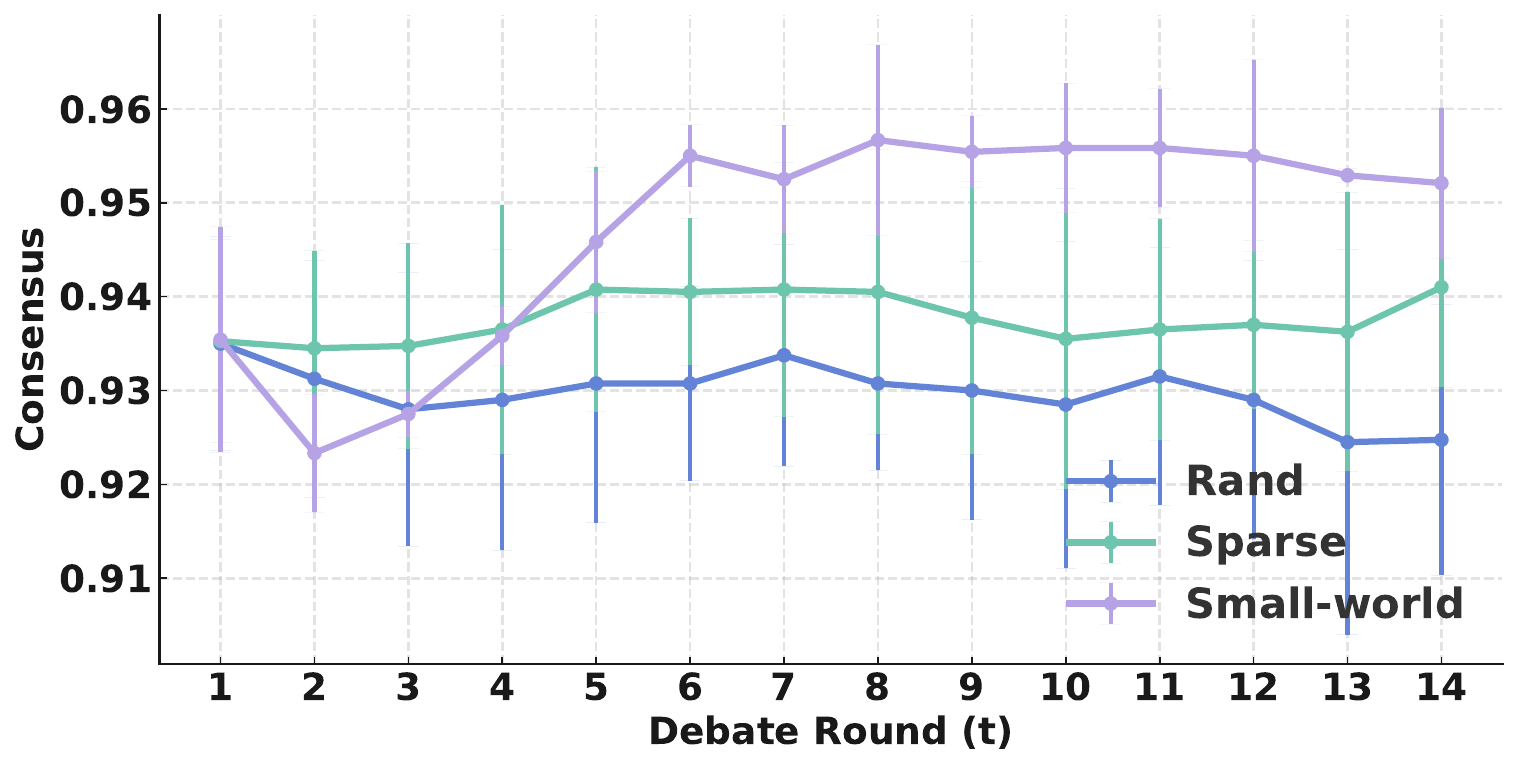}
    \caption{
        \textbf{Consensus dynamics on GSM8K using three communication topologies.}
        We compare \textit{Rand}, \textit{Sparse}, and \textit{Small-world} connectivity
        under the Multi-Agent Debate framework. 
        Small-world exhibits smoother convergence and lower variance than any other baselines.
    }
    \label{fig:gsm8k-consensus-errorbar}
\end{figure}

\section{Case Study: Small-World Structures improve Multi-Agent Debate}

\paragraph{Motivation and Setup.}
We select MAD, a foundational form of agent collaboration, as a case study to investigate and analyze the role of SW structures within collaborative multi-agent mechanisms. We believe that small-world structures can serve as an enhancing component for multi-agent systems, offering a promising improvement in stability. To provide empirical grounding for this claim, we conducted a lightweight experiment that compares several topologies. The evaluation includes a fully connected topology, a sparse ring, a random topology in which all edges are re-sampled at every debate round (noted as Rand), and a static topology with SW structures constructed by rewiring a small fraction of the ring edges. Our evaluation covers three representative benchmarks: a multiple-choice question answering benchmark ARC-Challenge~\cite{boratko-etal-2018-systematic}, a multimodal reasoning benchmark ScienceQA~\cite{10.5555/3600270.3600452}, and a widely-used mathematical reasoning dataset GSM8K~\cite{cobbe2021training}.

\paragraph{Accuracy and Cost Analysis.}
The results, summarized in Figure~\ref{fig:mad_topologies}, reveal an informative pattern. The Rand topology achieves the highest accuracy and the lowest communication cost across multiple tasks. This is consistent with classical SW theory: as the rewiring probability increases, a regular lattice gradually evolves into a SW network and eventually into a random graph. Within this continuum, the SW topology naturally occupies an intermediate region and therefore shows moderate accuracy at a moderate cost.

\paragraph{Consensus Dynamics.}
However, when we examine the consensus dynamics rather than only the answer-level outcomes, a different effect becomes visible, as shown by the curves presented in Figure~\ref{fig:gsm8k-consensus-errorbar}. The dynamic and highly sparse structure of the Rand topology makes the spread of correct intermediate reasoning signals irregular and often unstable. This effect is clearly visible in the consensus curves: the trajectories exhibit oscillations and, in some cases, even degrade over time, indicating that local errors can be amplified and propagated unpredictably across rounds. In contrast, the presence of structured long-range shortcuts in SW topology substantially stabilizes the consensus dynamics. 

\paragraph{Summary.}
It is important to note that in the context of MAD, the topology with SW structures sacrifices only a very small amount of additional communication cost while achieving accuracy that is nearly identical to the best-performing baselines. At the same time, it offers substantially more stable consensus trajectories. These observations highlight a crucial property of SW-guided communication: the topology not only preserves the benefits of local interaction patterns but also injects just enough long-range structure to prevent error accumulation and oscillatory behavior.

\paragraph{Future Research Opportunities.}
Future research may focus on the following directions suggested by our findings:  

\begin{itemize}[left=5pt]
\item  Analyzing how the placement and density of shortcut links influence consensus stability in multi-agent debate and other collaboration scenarios;

\item Identifying SW configurations that yield the most reliable convergence across divergent tasks.

\item Exploring dynamic mechanisms for generating and pruning the SW shortcuts in order to further improve the efficiency.
\end{itemize}

\begin{figure}[t]
    \centering
    \includegraphics[width=1\linewidth]{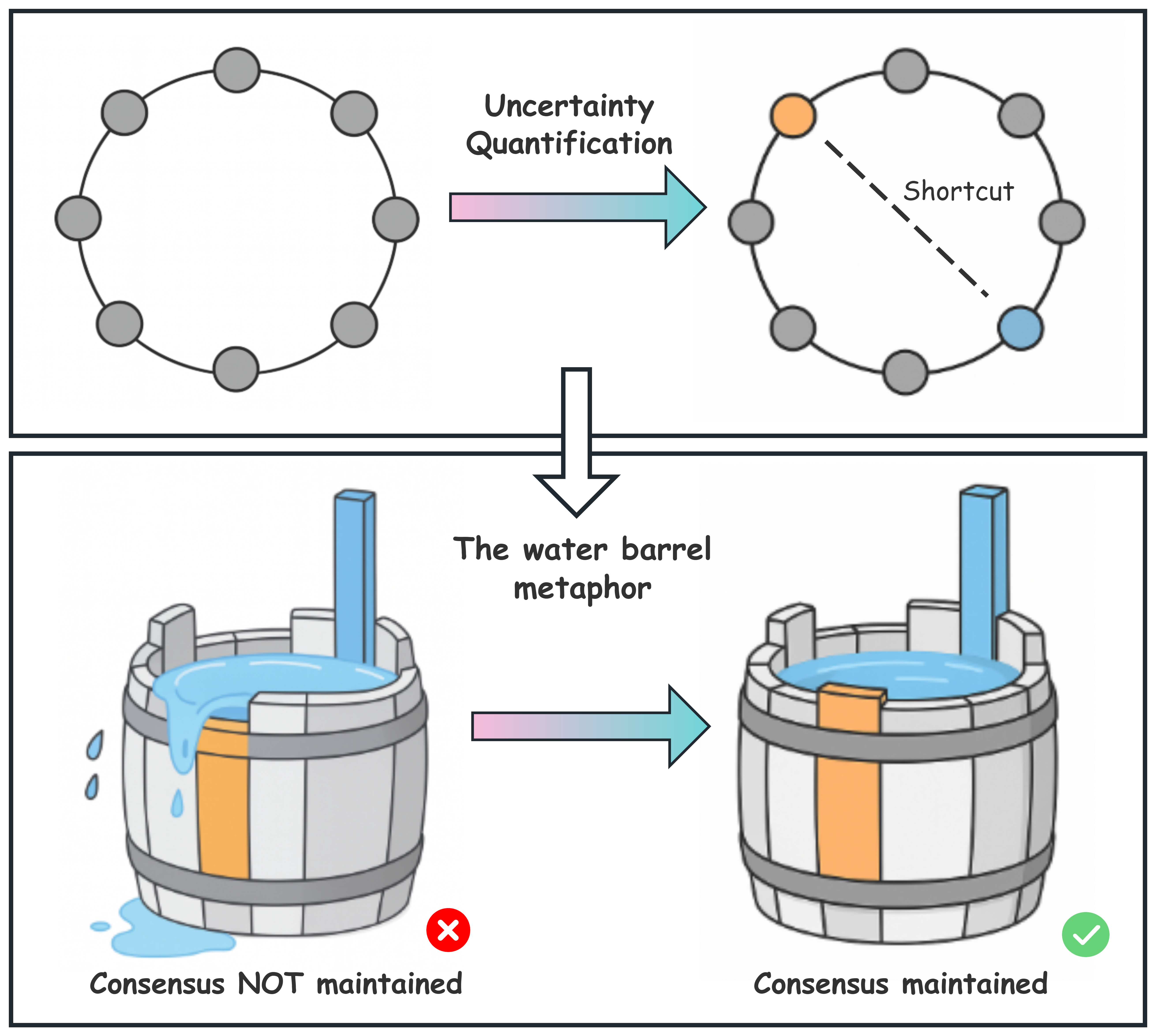}
    \caption{
Illustration of using uncertainty quantification (UQ) to create long-range shortcuts in a sparse multi-agent communication topology. Top: Ring topology with identified most uncertain and most confident agents, connected by a shortcut edge. Bottom: The water barrel metaphor: without the shortcut (left), consensus fails due to bottlenecks; with the shortcut (right), consensus is stabilized by improved information flow between epistemically divergent agents.
    }
    \label{fig:water}
    \vspace{-1em}
\end{figure}

\section{Toward Controllable Small-World Structures in Scaling Systems}

While our initial case study illustrates how SW structures stabilizes consensus formation in MAD, the broader implications of such structures extend beyond simple scenarios. In complex multi-agent coordination tasks such as distributed task allocation and collective exploration, recent studies suggest that shortcut-like communication patterns can either \emph{emerge naturally} or be deliberately constructed to enhance efficiency. 

\paragraph{Emergence in Complex Coordination.} 
For example, \citet{qian2025scaling} shows that when scaling LLM-based collaborative agents, small-world–like shortcuts arise spontaneously within learned communication policies. This indicates that such structures are not merely engineered artifacts but emergent organizational principles that LLM agents adopt to balance local coordination with efficient long-range information flow. \citet{Lou2024TAPE} further proposes incorporating agent topology into the policy gradient process to promote structure-aware communication. Other research demonstrates that sparse or hierarchical communication architectures can support scalable and effective behavior \citep{Liu2023DHCG, Niu2021MAGIC}, while graph attention mechanisms provide a flexible approach for capturing long-range dependencies in MAS \citep{Hu2024CommFormer}.

These findings resonate with broader insights from complex MAS, where small-world structures are linked to adaptability, robustness, and scalable decision-making. We argue that future MAS research should explore not only emergent structures, but also methods for \emph{actively constructing} such communication topologies. 

\paragraph{Uncertainty-Guided Shortcut Formation.} 
The key to forming a SW network is the induction of shortcut edges, which reduce the average path length and enhance the efficiency of information flow. We propose that a promising direction is to use \emph{uncertainty quantification} to guide the addition of short-cuts to form SW structures. For instance, \emph{semantic entropy} introduces a semantically grounded measure of uncertainty for LLMs, reflecting how widely the model’s predicted meanings diverge across plausible responses \cite{farquhar2024detecting, wang2025chasingconsistencyquantifyingoptimizing,han2024semantic}. With the aid of uncertainty quantification, agents holding divergent or low-confidence beliefs can be bridged to more confident agents. As illustrated in Figure \ref{fig:water}, we use an analogy to the classical “barrel effect,” where the overall capacity is determined by the shortest plank. In the context of multi-agent systems, small-world structures can be understood as introducing long-range shortcut edges that “compensate for the shortest plank” by directly linking weak or uncertain regions to stronger ones. This metaphor provides an intuitive criterion for when such edges are needed: long-range connections should be added precisely where agents exhibit the greatest divergence or lowest confidence. Using uncertainty quantification, we can identify the pair of agents with the most pronounced discrepancy in belief and proactively establish a shortcut between them. In this way, uncertainty-driven rewiring enables the deliberate formation of SW shortcuts,  facilitating more rapid convergence.

\paragraph{Design Implications.} 
This approach offers a pathway that is firmly rooted in theory for eliciting small-world properties in large-scale MAS. Rather than relying purely on self-organization, systems can reshape their topology in response to task demands, agent heterogeneity, or local uncertainty. We envision a new generation of multi-agent systems that blend adaptive communication with principled structural priors—forming sparse, shortcut-enriched networks that remain efficient, resilient, and cognitively diverse.

\paragraph{Future Research Opportunities.} 
Building on these insights, several promising research directions emerge. First, future work could investigate algorithmic frameworks for real-time topology adaptation using uncertainty signals, exploring how different uncertainty metrics (e.g., epistemic vs. aleatoric) influence shortcut formation. Second, there is an opportunity to develop optimization objectives that balance communication sparsity with semantic alignment, enabling more targeted and interpretable rewiring. Third, empirical studies are needed to understand how uncertainty-driven small-world topologies impact convergence speed, robustness, and generalization across diverse task regimes and agent configurations.

\section{Summary and Broader Implications}

In this section, we outline a broader set of implications driven by SW connectivity. We argue that SW structures support multi-agent systems in four critical dimensions: reasoning stability, robustness under uncertainty and perturbation, scalability in coordination, and the emergence of diverse cognitive roles.

\subsection{SW as Reasoning Stabilizer}
SW connectivity, which combines dense local clustering with occasional long-range shortcuts, naturally promotes stability over long decision horizons by enhancing information mixing and providing anchors against drift. In SW-structured MAS, local clusters function as cohesive sub-policies that help maintain consistency, while sparse shortcuts enable rapid propagation of global corrections. This effect has been observed in structured communication architectures. For example, engineered communication intention models using compact world models can significantly improve stability and sample efficiency in complex tasks compared to unstructured approaches~\cite{Hill2025}. Similarly, model-based sparse-communication schemes allow agents to transmit only when prediction errors increase, thereby correcting drift only when necessary~\cite{Han2023}.

\subsection{SW as Robustness Enhancement}
Sparse long-range SW links can act as backchannels that intercept or override local misinformation. By connecting distant clusters, SW shortcuts facilitate the diffusion of global consensus or corrections throughout the network. Recent work on adversarially robust MAS communication has leveraged this idea. For instance, the Graph Information Bottleneck framework compresses and filters messages to enhance resilience under noise and adversarial attacks~\cite{Ding2024}. In parallel, certifiably robust multi agent reinforcement learning (MARL) protocols employ ensemble-based decision processes to ensure that a corrupted message does not dominate the final outcome~\cite{Sun2023}. These techniques demonstrate how long-range connections can serve as anchors for accurate reasoning.

\subsection{SW as Scalable Coordinator}
SW-style topologies reduce communication and computational overhead by limiting interactions to local neighborhoods while augmenting them with sparse global links. Graph-aware protocols such as CommFormer are capable of selecting these sparse graphs automatically, achieving near full-information performance with significantly fewer messages~\cite{Hu2024}. Similarly, intention-based predictive models promote scalable behavior in large environments, improving both coordination and sample efficiency~\cite{Hill2025}. These structures enable most reasoning to occur locally while preserving the capacity for global coordination with minimal resource use~\cite{Han2023}.

\begin{figure}[t]
    \centering
    \includegraphics[width=0.8\linewidth]{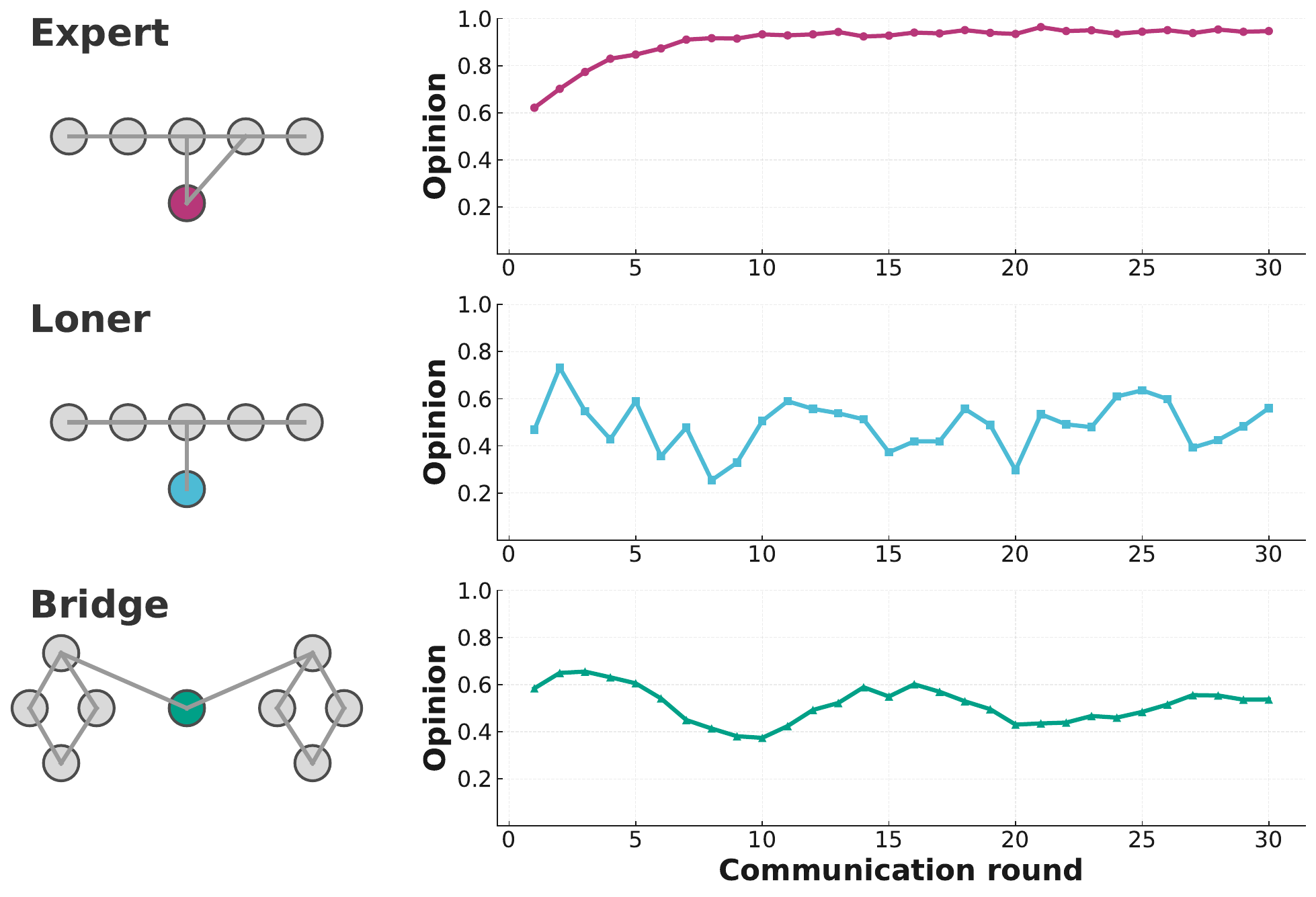}
    \caption{
Agents in different topological positions: Expert, Loner, and Bridge, showing distinct dynamics. SW structure induces diverse behaviors without explicit role design.
    }
    \label{fig:persona}
    \vspace{-1.3em}
\end{figure}

\subsection{SW as Inductive Bias for Cognitive Diversity}
SW networks naturally induce heterogeneous structural roles. These include 
densely connected hubs, cross community bridges, and sparsely connected peripheral nodes. 
In contrast to hand crafted prompting or explicit role engineering \cite{yao2023react,bai2022constitutionalaiharmlessnessai,wang2024voyager}, such asymmetries arise 
directly from the topology and provide an inherent source of behavioral diversity. Studies 
in role based MARL show that structural heterogeneity 
can substantially enhance coordination and robustness \citep{wang2020roma}. In a similar 
spirit, SW informed multi agent systems introduce a structural prior that encourages the 
emergence of differentiated cognitive roles. These roles enrich collective reasoning and more expressive dynamics.

To illustrate this idea, we conducted a simple simulation examining how an agent's 
topological position shapes its belief updating behavior. As shown in 
Figure~\ref{fig:persona}, we analyze three archetypal roles that naturally occur in SW 
topologies. The first is the \emph{Expert}, which corresponds to a hub like agent embedded in a 
tightly clustered neighborhood. The second is the \emph{Loner}, which corresponds to a weakly 
connected peripheral agent. The third is the \emph{Bridge}, which corresponds to an agent positioned 
on a shortcut edge that links two communities that would otherwise remain separate. These 
structural patterns emerge as the SW rewiring probability \(p\) changes.

The resulting opinion trajectories reveal clear behavioral differentiation. The Expert 
converges rapidly to a stable and high confidence belief due to strong local reinforcement. 
The Loner displays large fluctuations and instability because it receives weak and 
inconsistent signals. The Bridge evolves more slowly but integrates information from 
multiple communities. It often maintains intermediate or oscillatory beliefs as it 
mediates between competing local biases. This preliminary analysis demonstrates that even 
simple SW configurations can induce meaningful cognitive variation. Structural topology 
alone is sufficient to shape emergent behavioral roles in multi agent systems, without the 
need for explicit prompt design or manually specified roles.

\section{Conclusion and Outlook}

This paper revisits LLM-based MAS through the structural perspective of SW networks. Using MAD as a controlled testbed, we show that introducing SW shortcuts yields accuracy and communication cost comparable to strong baselines, while substantially stabilizing consensus trajectories. These findings suggest that SW connectivity provides a lightweight yet effective inductive bias for improving collective reasoning. Guided by this insight, we outline an uncertainty-driven rewiring scheme that adds shortcuts between epistemically divergent agents, enabling controllable and adaptive SW structures. Beyond MAD, such mechanisms may benefit broader multi-agent settings by balancing local specialization with efficient global information flow. 

Future research may explore how SW-inspired communication can be learned end-to-end, enabling agents to autonomously adapt their topology during interaction. Another promising direction is to generalize uncertainty-driven rewiring to heterogeneous or hierarchical agent populations. More broadly, studying how topology and reasoning co-evolve may provide deeper insights into scalable collective intelligence and cognitive diversity.


\bibliographystyle{ACM-Reference-Format} 
\bibliography{sample-base}


\end{document}